\pdfoutput=1
\documentclass[11pt]{article}

\usepackage[]{coling}

\usepackage{times}
\usepackage{latexsym}
\usepackage{booktabs}
\usepackage{multirow}
\usepackage{adjustbox}

\usepackage{enumitem}
\usepackage{times}
\usepackage{latexsym}
\usepackage{diagbox}

\usepackage[T1]{fontenc}
\usepackage[utf8]{inputenc}
\usepackage{hyperref}

\usepackage{microtype}
\usepackage{colortbl} 

\usepackage{inconsolata}
\usepackage{multirow}
\usepackage{graphicx}
\usepackage{amsmath}

\usepackage{amssymb}
\usepackage{xcolor}
\usepackage{booktabs}
\usepackage{arydshln}
\usepackage{subcaption}
\usepackage{needspace}
\usepackage{hyperref}

\definecolor{purple_f}{HTML}{B9B8EF}
\definecolor{purple_b}{HTML}{D4D5FF}
\definecolor{green_f}{HTML}{385723}
\definecolor{green_b}{HTML}{E2F0D9}
\usepackage[most]{tcolorbox}
\tcbset{on line, 
    boxsep=1.0pt, left=0pt, right=0pt,top=0pt,bottom=0pt,
    colframe=white, arc=1pt
}

\usepackage[most]{tcolorbox}

\usepackage[T1]{fontenc}

\usepackage[utf8]{inputenc}

\usepackage{microtype}

\usepackage{inconsolata}

\usepackage{graphicx}

\definecolor{arrowgreen}{RGB}{34, 139, 34}
\definecolor{arrowred}{RGB}{178, 34, 34}

\newcommand{\uparr}{\textcolor{arrowred}{$\uparrow$}}

\newcommand{\downarr}{\textcolor{arrowgreen}{$\downarrow$}}

\title{Unveiling Entity-Level Unlearning for Large Language Models: \\ A Comprehensive Analysis}
\author{
    Weitao Ma$^{\clubsuit}$ \quad Xiaocheng Feng$^{\clubsuit \spadesuit}$\thanks{$\quad$ means Corresponding Author} \quad Weihong Zhong$^{\clubsuit}$ \quad Lei Huang$^{\clubsuit}$ 
    \\
    \textbf{Yangfan Ye$^{\clubsuit}$ \quad Xiachong Feng$^{\diamondsuit}$ \quad Bing Qin$^{\clubsuit \spadesuit}$}\\
    $^{\clubsuit}$ Harbin Institute of Technology \\
    $^{\spadesuit}$ Peng Cheng Laboratory \\
    $^{\diamondsuit}$ The University of Hong Kong \\
    \texttt{\{wtma,xcfeng,whzhong,lhuang,yfye,qinb\}@ir.hit.edu.cn}, \texttt{fengxc@hku.hk}
}

\begin{document}
\maketitle

\begin{abstract}
Large language model unlearning has garnered increasing attention due to its potential to address security and privacy concerns, leading to extensive research in the field.
However, existing studies have predominantly focused on instance-level unlearning, specifically targeting the removal of predefined instances containing sensitive content.
This focus has left a gap in the exploration of removing an entire entity, which is critical in real-world scenarios such as copyright protection.
To close this gap, we propose a novel task named \textit{Entity-level unlearning}, which aims to erase entity-related knowledge from the target model completely. 
To investigate this task, we systematically evaluate popular unlearning algorithms, revealing that current methods struggle to achieve effective entity-level unlearning.
Then, we further explore the factors that influence the performance of unlearning algorithms, identifying that the knowledge coverage of the forget set and its size play pivotal roles. 
Notably, our analysis also uncovers that entities introduced through fine-tuning are more vulnerable than pre-trained entities during unlearning. 
We hope these findings can inspire future improvements in entity-level unlearning for LLMs.
\end{abstract}

\section{Introduction}{\label{sec:intro}}
Large Language Models (LLMs) \citep{achiam2023gpt, touvron2023llama, touvron2023llama2, meta2024introducing} pre-trained on extensive corpora have achieved significant success in various downstream tasks \citep{kamalloo2023evaluating, seegmiller2024llms}. 
However, training data often contains undesirable information, such as toxic texts \citep{lu2022quark}, privacy content \citep{liu2024model} and copyrighted information \citep{karamolegkou2023copyright}. These issues raise security and legal concerns, hindering the practical application of LLMs \citep{yao2024survey, das2024security}.
To tackle this, Machine Unlearning \citep{zhang2023right, lu2024eraser, bhardwaj2024language} has gradually been applied to LLMs due to its effectiveness and cost-efficiency. These refined techniques, now known as \textit{LLM Unlearning} \citep{yao2023large, liu2024rethinking, liu2024towards}, have become a mainstream approach for removing undesirable knowledge from the model by applying post-hoc modifications to target models. 

\begin{figure}[t]
\centering
\includegraphics[width=1.0\linewidth]{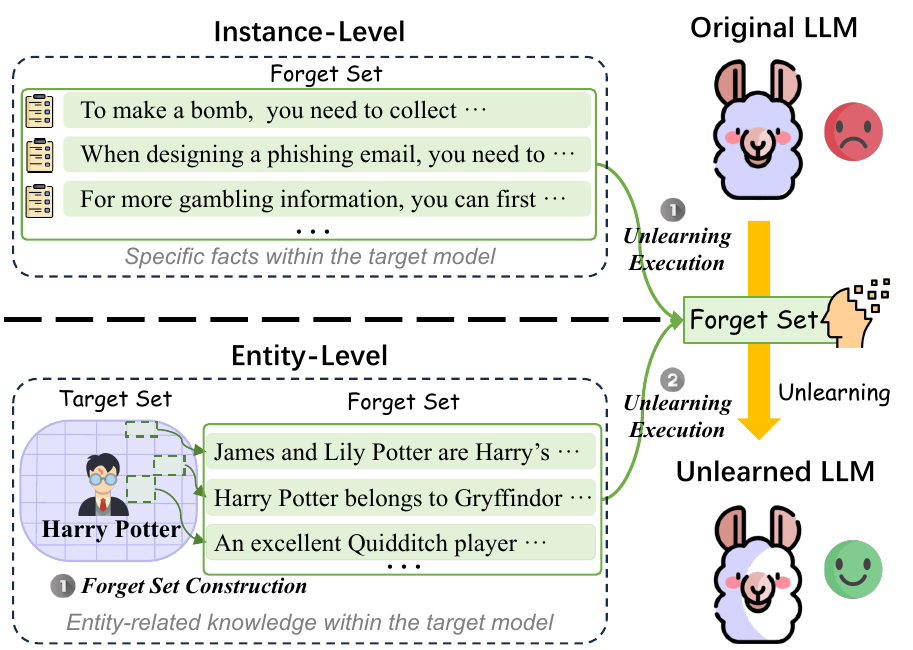}
\caption{The comparison between the instance-level unlearning process and entity-level unlearning process. The \tcbox[colback=purple_f]{knowledge} covered by the purple background represents the \textbf{target set}, and the \tcbox[colback=green_b]{knowledge} covered by the green background represents the \textbf{forget set}.
}
\label{fig:entity_instance_level}
\end{figure}

Towards this direction, the current unlearning paradigm involves applying an algorithm on a \textbf{forget set} which contains the undesirable knowledge \citep{zhang2024negative, maini2024tofu}. 
However, most existing research focuses on \textit{Instance-level Unlearning} tasks, which address isolated sensitive content \citep{li2024wmdp, ji2024beavertails}, while overlooking the deletion of entire entities, which is crucial in many real-world scenarios, such as removing `Harry Potter' for copyright protection \cite{eldan2024whos}.
To address this gap, we formally define a novel task named \textit{Entity-level Unlearning}.
As illustrated in Figure \ref{fig:entity_instance_level}, this task represents a significant divergence from instance-level unlearning due to the different nature of the information being targeted for removal.
Specifically, instance-level unlearning focuses on removing specific facts, which are predefined and can be directly erased. 
In contrast, entity-level unlearning addresses the removal of an entire entity, requiring the deletion of all knowledge associated with the entity within the model, which is referred to as the \textbf{target set}.
However, in real-world scenarios, the knowledge about an entity in the model is an abstract set that cannot be predefined.
Consequently, entity-level unlearning requires an additional step to construct a specific forget set containing the entity-related knowledge for deletion.

To this end, we propose a two-stage framework for the task, comprising \textit{forget set construction} and \textit{unlearning execution}. Specifically, in the first stage, we introduce a knowledge-probing method to obtain entity-related knowledge as the forget set. In the second stage, we perform removing the target entity by applying an unlearning algorithm to the constructed forget set.
To establish a controlled experimental environment, we choose to inject pseudo-entity knowledge into the model through the post-training, ensuring that \textit{all the knowledge of the target entity} (target set) is accessible. This allows us to assess the unlearning effect of pseudo-entity on the target set, which can achieve a precise evaluation.
Based on the settings, we systematically analyze the performance of five popular unlearning algorithms on the entity-level unlearning task within the two-stage framework and further investigate the pivot factors influencing the unlearning effect. Through extensive experiments, we have drawn several insightful conclusions:
\begin{itemize}[noitemsep,topsep=1pt,parsep=0pt,leftmargin=*]
\item Existing unlearning algorithms, which only focus on the forget set itself, struggle to achieve effective entity-level unlearning under real scenarios, making it necessary to develop corresponding unlearning algorithms that can generalize deletion effects to the entire entity's knowledge.
\item Increasing the knowledge overlap between the constructed forget set and the target set can enhance the entity deletion effect of algorithms. However, simply expanding the size of the set is an ineffective way to improve this overlap, which will destroy the general ability of the model.
\item The entities introduced through fine-tuning are more vulnerable than the pre-trained entities during unlearning, highlighting the need for more robust methods for injecting target knowledge to simulate pre-trained knowledge better.
\end{itemize}

In summary, removing an entire entity from LLMs is a realistic and challenging task. We encourage future research to build on our findings by exploring more precise knowledge-probing methods and developing targeted unlearning algorithms to enhance the effectiveness of entity deletion.

\section{Entity-level Unlearning}\label{sec:Defination}
\subsection{Task Definition and Setting}
The objective of the entity-level unlearning task is to remove an entire entity from the target model. However, the knowledge associated with an entity is an abstract concept, which cannot be predefined in real-world scenarios. 
Therefore, it is necessary to construct a specific set that contains the model's knowledge of the entity for deletion.
Consequently, we define this task as \textit{removing the entire entity from the target model by deleting a specific set containing entity-associated knowledge}.

The entity-level unlearning task can be formalized as follows: Given a target entity \textit{O}, the target model parametrized by $\theta_t$ is required to forget a target set $S_T$, which contains all knowledge related to the entity \textit{O} in the model, by applying unlearning methods $\textsc{H}(\cdot)$ on a forget set $S_F$, which contains the specific knowledge of the entity. The unlearning process can be expressed as follows:
\begin{equation}
    \theta_{t+1} \leftarrow \textsc{H}(\theta_t, S_F)
\end{equation}

To precisely assess the deletion effect of the target entity, the evaluation for entity-level unlearning task $\textsc{E}(\cdot)$ should be conducted on the target set $S_T$:
\begin{equation}
    Score_{forget} = \textsc{E}(\theta_{t+1}, S_T)
\end{equation}

A significant challenge in evaluating the effectiveness of entity-level unlearning posed by the inability to access all their training data of LLMs is obtaining the target set $S_T$ used to evaluate the unlearned models. To address this, we simulate entity-level unlearning scenarios following the TOFU \citep{maini2024tofu}, which fine-tunes the LLMs using a fictitious author dataset. The dataset ensures that the LLMs have no prior exposure to these authors during previous training phases. Thus, the fine-tuning dataset encompasses all the knowledge about the entities, making it suitable to serve as the target set $S_T$.

\subsection{Entity-level Unlearning Framework}
In light of the definition of the entity-level unlearning task, we propose a two-stage framework for the task, consisting of: 1) \textit{Forget Set Construction} and 2) \textit{Unlearning Execution}. Building on this framework, we design entity-level unlearning methods based on trending unlearning algorithms.

\subsubsection{Forget Set Construction}
\label{forget_constrcut}
The first step of the entity-level unlearning task is constructing a forget set, which is both critical and challenging. In practical scenarios, the forget set often would be generated solely based on the target model and the names of the entities involved. Following this setting, we propose a simple and effective constructing baseline to probe the entity-related question-answer (QA) pairs within the target model as a forget set. More details can be found in Appendix \ref{app_framework}.

Specifically, we first prompt the target models to self-generate entity-related questions according to their internal knowledge, inspired by \citet{weller2023according}. Note that only those non-repetitive questions containing the entities' names will be kept to further ensure the focus remains on the corresponding entities. Next, we acquire the answers from the target model using greedy decoding, which ensures that the answers possess a relatively high generation probability. Finally, we conduct a self-verification process, where the model repeatedly evaluates each QA pair, retaining only the pairs for which the model consistently agrees with its responses. These selected pairs serve as valid candidates for forget sets of unlearning algorithms.

This approach offers a straightforward knowledge extraction method for acquiring the forget set. We then assess the influence of the forget set's quality on unlearning methods through a manual replacement analysis.

\subsubsection{Unlearning Execution}
After constructing the forget set, the next step is to apply the unlearning algorithms to it. In the absence of algorithms specifically designed for entity-level unlearning, we select five representative algorithms for evaluation. One of the most straightforward unlearning methods is Gradient Ascent, which reduces the likelihood of the answers to achieve unlearning. Other methods introduce additional constraints on a retain set, which contains knowledge that should be preserved to minimize the damage to the model. An ideal entity-level unlearning method should effectively remove the entire entity while minimizing any negative impact on the remaining knowledge. This study primarily examines the performance of current unlearning algorithms on the entity-level unlearning task, without yet delving into the broader algorithmic application framework \citep{huang2024offset}.

\section{Experiments}\label{sec:experiments}
\subsection{Experimental Setup}
\paragraph{Datasets and models.} We conduct entity-level unlearning experiments on the TOFU benchmark \citep{maini2024tofu}, which includes synthetically generated biographies of 200 fictitious authors, each consisting of 20 question-answer pairs, under some new experimental settings. We fine-tune the Llama2-7B-Chat \citep{touvron2023llama2} and Phi-1.5 \citep{li2023textbooks} on the TOFU dataset as the target models. See the Appendix \ref{target_model} for details. Additionally, we also construct the \textit{target set}, \textit{forget set}, \textit{retain set}, and \textit{evaluation set} required for the experiment. The dataset collection and composition are as follows (more details can be found in Appendix \ref{app_eval_set}): 
\begin{itemize}[noitemsep,topsep=0pt,parsep=1pt,leftmargin=*]
\item \textbf{Target Set}: For the target entity, we select the oracle training dataset with 20 question-answer pairs in TOFU as the target set. 
\item \textbf{Forget Set}: 
In the experiments, we define two types of forget sets. The first type involves selecting the \textit{target set} as the forget set for the unlearning algorithms, simulating an ideal scenario. The second type is constructed by generating a \textit{constructed forget set (Con. forget set)}  from the target model, following the methods outlined in Section \ref{forget_constrcut}. The constructed forget set consists of 20 QA pairs related to the target entity.
\item \textbf{Retain Set}: The retain set is constructed from TriviaQA \cite{joshi2017triviaqa} and consists of QA pairs related to world knowledge. We ensure that each question is correctly answered by both two target models using greedy decoding.
\item \textbf{Evaluation Set}: We assess the unlearned model on the evaluation set, including the target set, the retain set, the real authors set, and the world facts set. The latter two datasets, sourced from TOFU, are used to evaluate the retention of pre-training knowledge in unlearned models. 
\end{itemize}

\begin{table*}[!ht]
\centering
\resizebox{\textwidth}{!}{
\begin{tabular}{llcccccccc}
\toprule
\multirow{2}{*}{\textbf{Method}} & \multirow{2}{*}{\textbf{Forget Set Type}} & \multicolumn{4}{c}{\textbf{Target Set}} & \multirow{2}{*}{\textbf{RS Score} {\uparr}} & \multirow{2}{*}{\textbf{RAS Score} {\uparr}} & \multirow{2}{*}{\textbf{WFS Score} {\uparr}} & \multirow{2}{*}{\textbf{Model Utility} {\uparr}} \\
\cmidrule(lr){3-6}
&  & \textbf{Prob.} {\downarr} & \textbf{ROUGE} {\downarr} & \textbf{Acc.} {\downarr} & \textbf{Forget Q.} {\uparr} & & & & \\
\midrule 
\rowcolor{gray!20}\multicolumn{10}{c}{\textbf{Llama2-7B-Chat-TOFU}} \\ 
\midrule

Original & - & 0.9908 & 0.9793 & 0.655 & 0.0300 & 0.8737 & 0.5893 & 0.5308 & 0.6349  \\
\midrule

\multirow{2}{*}{Grad. Ascent} & target set & \textbf{0.0009} & 0.2319 & \textbf{0.2250} & \textbf{0.1740} & 0.6803 & 0.5533 & 0.5020 & \textbf{0.5694} \\
 & Con. forget set  & 0.0411 & \textbf{0.1858} & 0.4325 & 0.1012 & 0.5685 & 0.5391 & 0.4841 & 0.5282 \\
\midrule

\multirow{2}{*}{Grad. Diff.} & target set & 0.1237 & 0.3717 & \textbf{0.3950} & \textbf{0.3604} & 0.8326 & 0.6584 & 0.5899 & \textbf{0.6795} \\
 & Con. forget set  & \textbf{0.1142} & \textbf{0.2784} & 0.4450 & 0.1177 & 0.7842 & 0.4899 & 0.4710 & 0.5515 \\
\midrule

\multirow{2}{*}{KL Min.} & target set & \textbf{0.0002} & \textbf{0.1110} & \textbf{0.2125} & \textbf{0.2430} & 0.6203 & 0.5448 & 0.5104 & \textbf{0.5549} \\
 & Con. forget set  & 0.0435 & 0.1803 & 0.4250 & 0.1007 & 0.5765 & 0.5440 & 0.4841 & 0.5321 \\
\midrule

\multirow{2}{*}{Pref. Opt.} & target set & \textbf{0.3486} & \textbf{0.0147} & 0.5150 & \textbf{0.2981} & 0.9024 & 0.6777 & 0.6349 & 0.7213 \\
 & Con. forget set  & 0.3757 & 0.0382 & \textbf{0.5075} & 0.2765 & 0.9045 & 0.6986 & 0.6372 & \textbf{0.7306} \\
\midrule

\multirow{2}{*}{NPO-GD} & target set & \textbf{0.0344} & \textbf{0.2971} & \textbf{0.3225} & \textbf{0.5253} & 0.7887 & 0.5715 & 0.5286 & \textbf{0.6111} \\
 & Con. forget set  & 0.5076 & 0.5000 & 0.5300 & 0.0875 & 0.7464 & 0.4546 & 0.4616 & 0.5258 \\

\bottomrule 
\end{tabular}
}
\caption{
The performance of the Llama2-7B-Chat-TOFU after entity-level unlearning under the five algorithms. We list the results when the Forget Quality (Forget Q.) reaches the first peak score. Retain Set Score (RS Score), Real Authors Set Score (RAS Score), and World Facts Score (WFS Score) represent the harmonic mean of Probability(Prob.), ROUGE, and Accuracy(Acc.) on their respective sets. \uparr \ \ represents that the higher score is better, while \downarr \ \ indicates the opposite. The results in \textbf{bold} represent the best results between the two forget sets.
}
\label{table:main-results}
\end{table*}

\paragraph{Unlearning algorithms.} We experiment with five common unlearning algorithms on the entity-level unlearning task (more details can be found in Appendix \ref{app_baseline}): 
\begin{itemize}[noitemsep,topsep=1pt,parsep=0pt,leftmargin=*]
\item \textbf{Gradient Ascent (Grad. Ascent)} \citep{yao2023large}, which is one of the most straightforward unlearning methods, reduces the likelihood of original answers on the forget set $S_F$.
\item \textbf{Gradient Difference (Grad. Diff.)} \citep{liu2022continual}, which is a variant of Grad. Ascent, not only implements unlearning on the forget set $S_F$ but also learning on the retain set $S_R$ by gradient descent to minimize damage to the model.
\item \textbf{KL Minimization (KL Min.)} applies an additional Kullback-Leibler (KL) divergence regularization between the predictions on $S_R$ of the original model $\theta_t$ and the unlearned model $\theta_{t+1}$, while performing Grad. Ascent on $S_F$.
\item \textbf{Preference Optimization (Pref. Opt.)} optimizes the model to realign the questions in the forget set $S_F$ with a refusal answer, such as "I don't know," through the DPO algorithm \citep{rafailov2024direct} while learning on the retain set $S_R$ by gradient descent.
\item \textbf{Negative Preference Optimization} is an efficient unlearning method that requires only providing a negative response during preference optimization. We adopt the algorithm with the same restriction on the retain set $S_R$ as Gradient Difference (NPO-GD), which has been proven to outperform other variants \citep{zhang2024negative}.
\end{itemize}

\paragraph{Evaluation metrics.} We assess the unlearned models using the following metrics (more details can be found in Appendix \ref{app_mertic}):
\begin{itemize}[noitemsep,topsep=0pt,parsep=0pt,leftmargin=*]
\item \textbf{ROUGE} \citep{lin2004rouge} measures the overlap co-occurrence of n-grams between the original answer and the model's greed-decoding generation for the test QA pairs. 
\item \textbf{Probability} computes the conditional probability with length normalization of QA pairs in the evaluation set.
\item \textbf{Accuracy} calculates the proportion of a paraphrased answer that the unlearned model can select from perturbed answers of the question. 
\item \textbf{Forget Quality} assesses the unlearning effectiveness of the unlearned models via the Kolmogorov-Smirnov (KS) test. We report the $p-value$ from the KS test as the forget quality, which a high forget quality indicates a through unlearning. 
\end{itemize}
We assess the performances by evaluating ROUGE scores, probabilities, and accuracy across all evaluation sets. To derive a comprehensive measure of generative performance, we compute the harmonic mean of the nine values obtained from the retain set, real authors set, and world facts set, referring to this as \textit{Model Utility}, in line with TOFU. Additionally, we evaluate the forget quality exclusively on the target set. 

\subsection{Experimental Results}
\label{experimental_results}
We present experimental results comparing the performance of the same method on various forget sets and between different algorithms. The main experimental results of Llama2-7B-Chat-TOFU are shown in Table \ref{table:main-results}, and the experimental results of Phi-1.5-TOFU are presented in Appendix \ref{experiments_results}.

A comparison of the performance across different types of forget sets, as presented in Table \ref{table:main-results}, reveals that the algorithms based on the target set consistently maintain similar model utility while achieving lower probability, reduced accuracy, and overall higher forget quality on Llama2-7B-Chat-TOFU. A similar trend is observed in the Phi-1.5-TOFU model, as shown in Table \ref{table:main-phi_results}. These results suggest that applying the unlearning algorithms to the target set results in more thorough forgetting than their application on the constructed forget set. This finding highlights that the current unlearning algorithms struggle to generalize effectively to entity-level unlearning tasks when relying on the constructed forget set. Consequently, the construction of the forget set is crucial in determining the success of entity-level unlearning. In the subsequent analysis, we will explore in detail how the quality of the forget set influences unlearning effectiveness.

Comparing the performance of different unlearning algorithms reveals several insights critical for enhancing entity-level unlearning:
\begin{enumerate}[leftmargin=*,itemsep=0em]
    \item The Grad. Ascent method can effectively minimize the probability of the ground truth answer but harms the model's ability, resulting in the lower forget quality and diminished model utility. Comparing its two variants, the KL. Min. and Grad. Diff. methods, learning through gradient descent on the retain set proves more effective than the KL restriction in remedying the excessive damage to the model.
    \item Although the Pref. Opt. method significantly reduces the ROUGE of the target set answers and achieves relatively high forget quality on both two target models, the original answers still maintain a high generation probability and accuracy. This suggests that the method performs unlearning by increasing the likelihood of refusal answers rather than truly forgetting the target entity.
    \item The NPO-GD method stands out as one of the strongest approaches in unlearning, particularly excelling in instance-level unlearning tasks under ideal conditions. However, it exhibits the poorest performance on the constructed forget set, with a significant disparity in forget quality between the target set and the constructed forget set: 0.5253 vs. 0.0875 for Llama2-7B-Chat-TOFU and 0.5104 vs. 0.0794 for Phi-1.5-TOFU, according to Table \ref{table:main-results} and \ref{table:main-phi_results}. This substantial disparity may be attributed to the current optimization focus of unlearning algorithms, which tends to prioritize the forget set itself, often in opposition to the goal of generalizing within the entity, a crucial aspect for effective entity-level unlearning. This highlights the necessity for developing a targeted algorithm tailored specifically for entity-level unlearning.
\end{enumerate}

\section{Analysis}\label{sec:analysis}
\begin{figure}[tp]
\centering
\includegraphics[width=1.0\columnwidth]{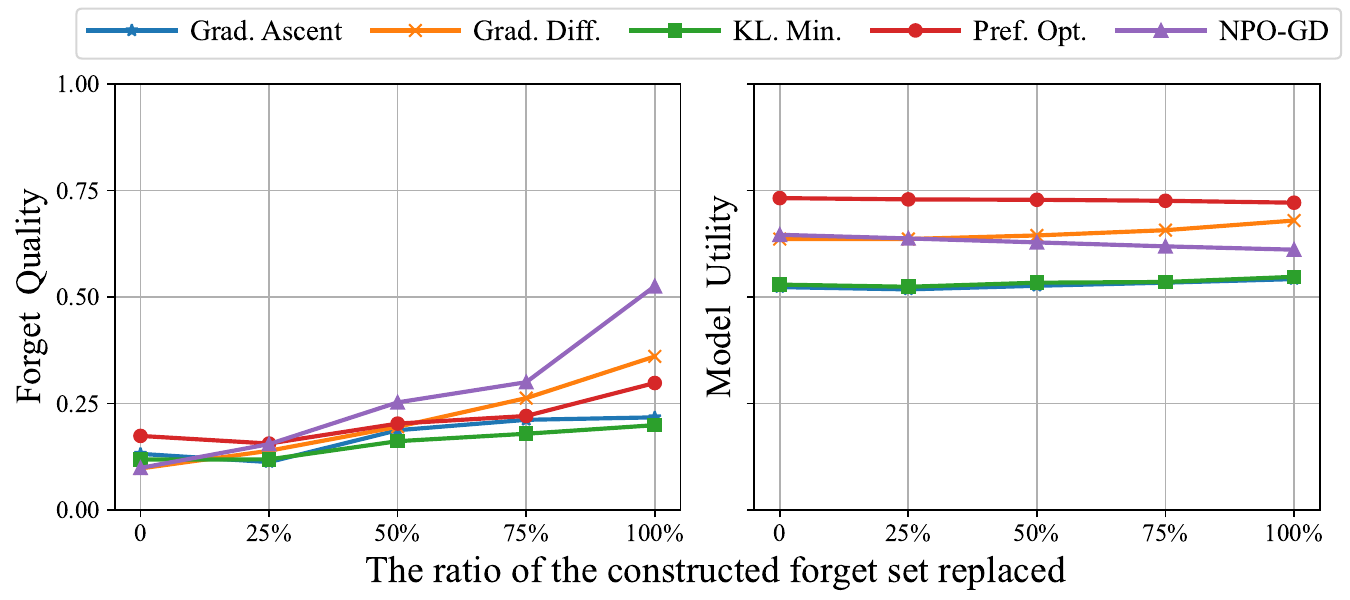} 
\caption{The performance of five unlearning algorithms in forget quality and model utility metrics across different constructed forget sets, achieved by replacing varying ratios of QA pairs
from the target set.}
\label{quality_ablation}
\end{figure}

The main experimental results indicate that the unlearning algorithms exhibit limited effectiveness in removing entities with the constructed forget set. In this section, we analyze the pivot factors that may influence the effect of the algorithms. Our subsequent analysis is conducted on the Llama2-7B-Chat-TOFU model, which is widely used.

\subsection{Effect of Forget Set} 
We hypothesize that the limited unlearning effect is due to the insufficient coverage of knowledge within the current constructed forget set. To explore this, we introduce a new metric, \textit{Knowledge Coverage}, designed to assess the knowledge overlap between the constructed forget sets and the target set. This metric is computed using the BERTScore \cite{zhang2019bertscore} of the closest QA pair match between the constructed forget set and the target set. A higher knowledge coverage indicates that the constructed forget set encompasses a broader range of entity-related knowledge. See Appendix \ref{forget_set_appendix} for detailed definitions and formulas.

To further investigate the effects of varying knowledge coverage, we construct forget sets with varying degrees of coverage by systematically replacing different ratios of QA pairs within the constructed forget set with those from the target set while keeping the total set size fixed at 20. As demonstrated in Table \ref{tab:forget_set_quality}, the knowledge coverage of the constructed forget set increases progressively as the ratio of replaced pairs grows. Then, we applied the five algorithms to perform entity-level unlearning on the constructed forget set under five different replacement ratios. As illustrated in Figure \ref{quality_ablation}, enhancing knowledge coverage of the constructed forget set consistently enhances the forget quality of most methods. Furthermore, each algorithm consistently preserved similar model utility across various constructed forget sets. These results suggest that \textit{enhancing the knowledge coverage of the constructed forget set can improve the unlearning effectiveness}.

\begin{figure}[pt]
\centering
\includegraphics[width=1.0\columnwidth]{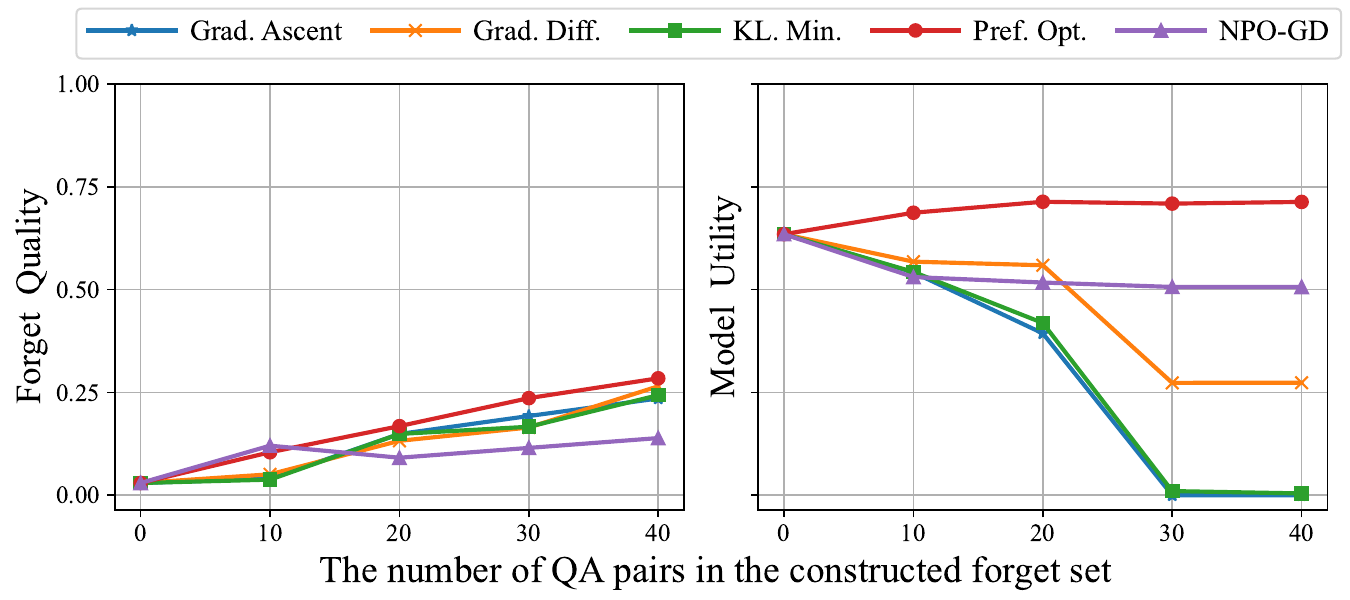} 
\caption{The performance of five unlearning algorithms on the constructed forget sets across different scales in forget quality and model utility metrics }
\label{quantity_ablations}
\end{figure}

An intuitive approach to increase knowledge coverage is by increasing the size of the constructed forget set. Therefore, we further explore the performance of the algorithms on constructed forget sets across different sizes. Specifically, we expand the constructed forget set size for each entity and evaluate the performance of the unlearning algorithms across five different sizes. As illustrated in Figure \ref{quantity_ablations}, the forget quality of the five unlearning algorithms gradually improves as the size of the constructed forget set increases. However, except for the Pref. Opt. method, the model utility of the unlearned models produced by the other algorithms demonstrates a noticeable decline. 
This decline in model utility, when compared to the trends in Figure \ref{quality_ablation}, is likely due to the larger number of unlearning steps required as the constructed forget set size grows. 
These findings suggest that, for most unlearning algorithms, \textit{expanding the constructed forget set size leads to a trade-off between forget quality and model utility}.
Furthermore, according to section \ref{experimental_results}, although the Pref. Opt. method maintains a high generation probability for the forget knowledge, it fails to achieve true deletion, as the forgotten information can still be easily retrieved through fill-in-the-blank questions \citep{wang2024rkld}. 

Furthermore, we also measure the knowledge coverage and record the probing rounds of constructed forget sets across different sizes. As shown in Table \ref{tab:forget_set_size}, knowledge coverage truly increases with the size of the constructed forget set, but the rate of increase gradually diminishes. Simultaneously, expanding the size of the constructed forget set proves to be challenging. We employ the knowledge-probing method to iteratively enlarge the dataset and report the number of iterations required for different set sizes in Table \ref{tab:forget_set_size}. The results indicate that once the constructed forget set reaches a certain size, further expansion of the constructed forget set becomes increasingly resource-intensive. The analysis mentioned indicates that simply increasing the size of the constructed forget sets does not effectively enhance the unlearning performance. 

\begin{figure*}[pt]
\centering
\includegraphics[width=1.0\textwidth]{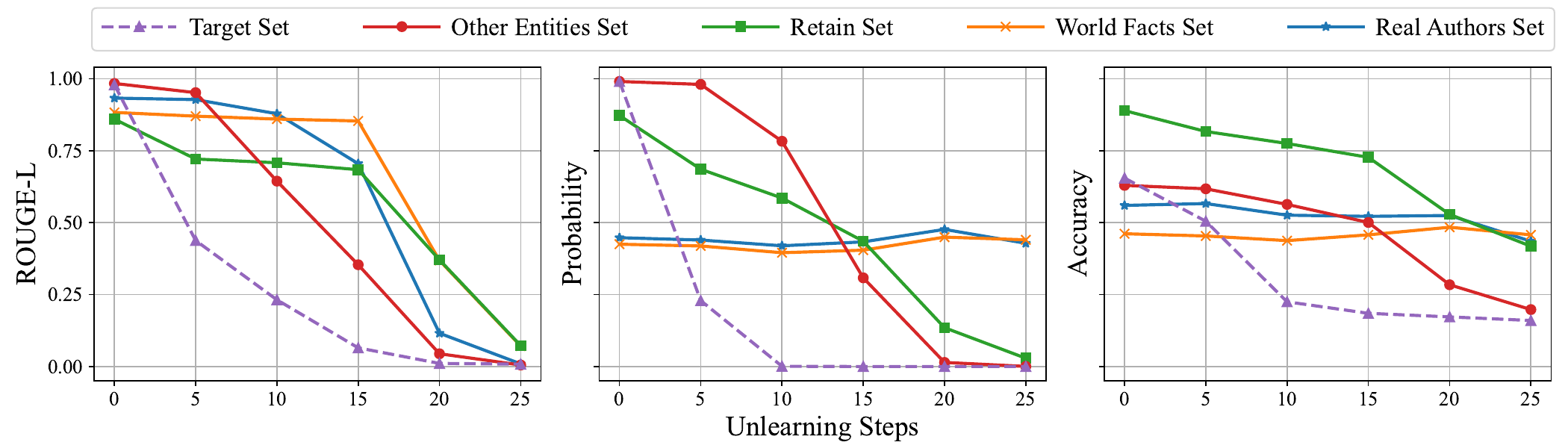} 
\caption{Step ablation analysis of unlearning Llama2-7B-Chat-TOFU using the Grad. Ascent algorithm. We report the ROUGE, probability, and accuracy for the evaluation sets at intervals of 5 steps, ranging from 0 to 25 steps.}
\label{llama2_step_GA}
\end{figure*}

\begin{table}[t!]
\small
    \centering
    \begin{tabular}{cll}
    \toprule
        \textbf{Size} & \textbf{Knowledge Coverage} & \textbf{Rounds}  \\
         \midrule
         10 &  0.2090 & 1.00 \\
         20 &  0.3074 (+0.0984) & 6.45 (+5.45) \\
         30 &  0.3656 (+0.0582) & 17.35 (+10.90) \\
         40 &  0.4146 (+0.0490) & 35.75 (+18.40) \\
         \bottomrule
    \end{tabular}
    \caption{Knowledge Coverage across various constructed forget set sizes and the average number of probing rounds required to obtain each set. Values in () indicate the increase relative to the previous set.}
    \label{tab:forget_set_size}
\end{table}

In summary, the quality of the constructed forget set plays a crucial role in the unlearning effect. A constructed forget set with greater knowledge coverage results in better unlearning performance. However, using the large constructed forget sets would compromise model utility and incur higher costs. Therefore, we encourage future research to explore more precise knowledge probing methods to construct higher-quality constructed forget sets.

\subsection{Effect of Unlearning Steps}
The number of unlearning steps, which determine the frequency of model parameter updates, plays a crucial role in the degree of change within the model. From our earlier analysis, we infer that unlearning steps significantly impact the model's utility. In this section, we evaluate the performance of the five algorithms on the target set across different unlearning steps. This evaluation is conducted using ROUGE, probability, and accuracy metrics following TOFU. Additionally, we examine the effect on fine-tuning knowledge during the unlearning process by sampling non-target entity knowledge, referred to as the \textit{Other Entities Set}.

As illustrated in Figure \ref{llama2_step_GA}, the Grad. Ascent method leads to a decline in ROUGE, probability, and accuracy across all sets as the number of unlearning steps increases. This trend aligns with the results observed in the performance of the KL. Min. in Figure \ref{llama2_step_kl}. Compared to the KL. Min. method, the Grad. Diff. algorithm in Figure \ref{llama2_step_GD} more effectively mitigates the negative impact on model performance, particularly on the \textit{retain set}, \textit{world facts set}, and \textit{real authors set}, corroborating the results discussed in Section \ref{experimental_results}.
Additionally, the Pref. Opt. algorithm in Figure \ref{llama2_step_idk} consistently maintains high probability and accuracy for the target set. This further suggests that the Pref. Opt. algorithm may not achieve true unlearning but increases the likelihood of refusal answers. In summary, while increasing the number of unlearning steps generally damages the model's overall capabilities, this damage can be mitigated by applying appropriate constraints on a \textit{retain set}, such as gradient descent.

Additionally, we observe that during unlearning, all algorithms show a more pronounced decline in performance on the \textit{other entities set} compared to the other non-target sets including the \textit{retain set}, \textit{world fact set} and \textit{real authors set}, which are introduced by pre-training. This phenomenon suggests that entity knowledge introduced through fine-tuning may be more susceptible during unlearning. 
Based on this observation, we further compare the performance of the \textit{other entities set} to that of the other non-target sets in both pre-training and fine-tuning scenarios.

\begin{figure*}[h]
\centering
\includegraphics[width=1.0\textwidth]{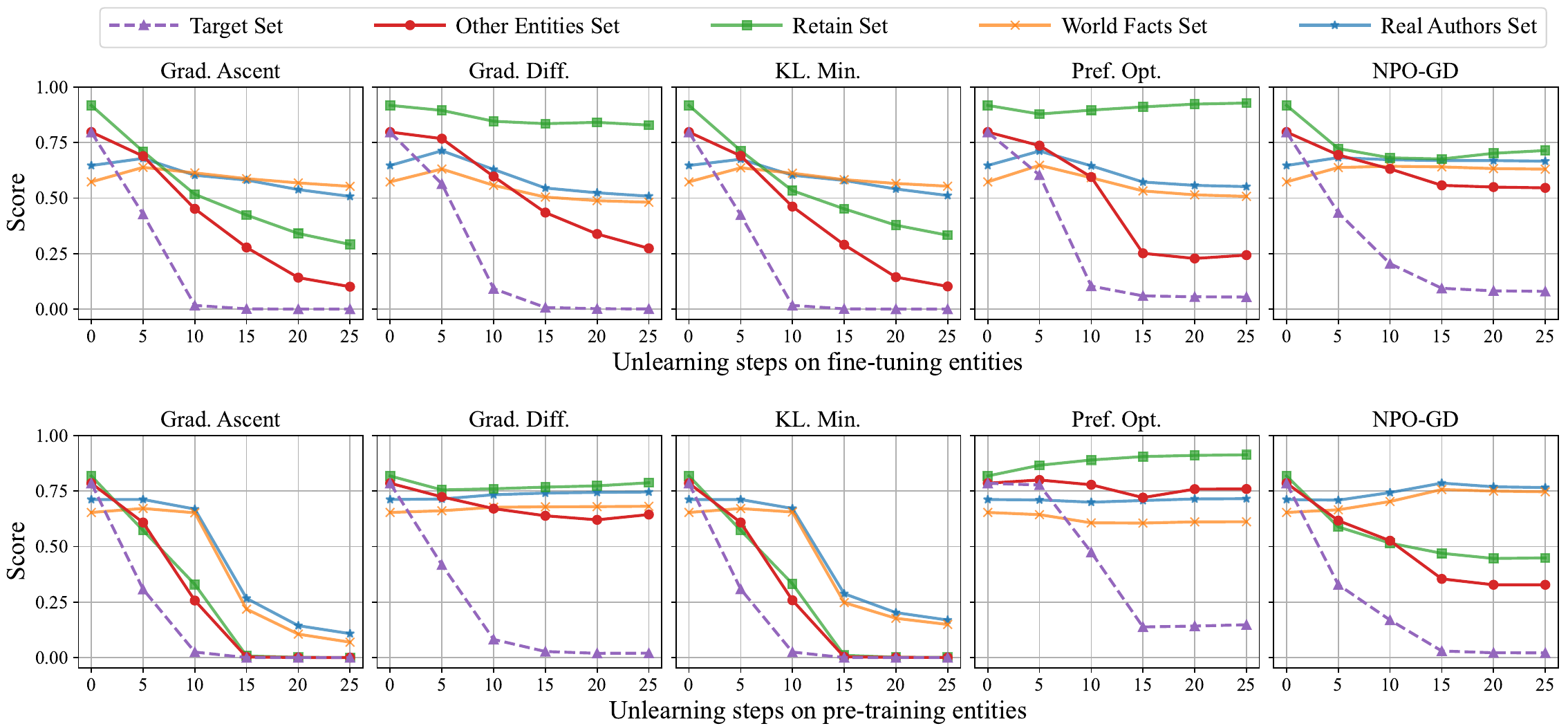} 
\caption{The comparison of the five algorithms during unlearning on both pre-trained and fine-tuned entities. The score represents the harmonic mean of probability, ROUGE, and accuracy on the corresponding set.}
\label{pretrain_fine-tuning_results}
\end{figure*}

\subsection{Entities learned in Pre-training v.s. Fine-tuning}
To establish a comprehensive comparison between pre-trained and fine-tuned entities, we select the Llama2-7B-Base model as the target model. Specifically, we fine-tune the model to inject a set of pseudo-entity knowledge and extract a set of celebrity knowledge introduced from pre-training as the target sets, while ensuring the same number of QA pairs for each entity. Subsequently, we perform entity-level unlearning under the two scenarios and assess the performance of the unlearned model using corresponding evaluation sets. More details can be found in Appendix \ref{pretraining_fintuning}.

We present the unlearning performance of the five algorithms on pre-trained and fine-tuned entities, as illustrated in Figure \ref{pretrain_fine-tuning_results}. For the pre-trained entities, all five methods display a consistent trend between the \textit{other entities set} and the other non-target sets. In contrast, for the fine-tuned entities, the Grad. Diff., KL Min., and Pref. Opt. algorithms cause more significant damage to the \textit{other entities set} than the other non-target sets. These findings suggest that the knowledge introduced during fine-tuning is more vulnerable to unlearning interventions, confirming our hypothesis about its fragility compared to the original pre-trained knowledge. Based on this analysis, future research should explore more robust methods for injecting target knowledge in entity-level unlearning tasks, aiming to simulate better the pre-trained knowledge which constitutes a significant majority of the knowledge within LLMs.

\section{Related Work}\label{sec:related_work}
\subsection{Algorithms of LLM Unlearning} 
LLM unlearning \citep{liu2024rethinking} has attracted rising attention owing to its potential to address privacy \citep{jang2022knowledge} and security \citep{barrett2023identifying} concerns.
In addition, the LLM unlearning method can effectively mitigate the hallucination problem \citep{huang2023survey} in LLMs and analyze the influence of the training knowledge \citep{zhao2024deciphering}.
Current unlearning algorithms for LLMs \citep{cao2015towards, yu2023unlearning, jang2022knowledge} focus on minimizing the impact of undesirable training data while preserving the integrity of other model knowledge based on the \textit{Forget Set} and \textit{Retain Set}. Specifically, these methods \citep{jang2022knowledge, chen2023unlearn} apply Gradient Ascent (GA) on the Forget Set to unlearn and add additional auxiliary loss on the Retain Set akin to gradient descent \citep{liu2022continual} and KL minimization \citep{maini2024tofu} to mitigate undesired effects. Additionally, inspired by the alignment capabilities of reinforcement learning from human feedback (RLHF) \citep{ouyang2022training, bai2022training}, several researchers adopt preference optimization methods to refine model outputs, like Direct Policy Optimization (DPO) \citep{rafailov2024direct} and Distributional Dispreference Optimization (D2O) \citep{duan2024negating}. Moreover, some researchers utilize model editing methods \citep{yao2023editing, feng2023trends, zhang2024comprehensive} to remove sensitive information by adjusting knowledge-related parameters \citep{wu2023depn, wang2024detoxifying}. However, it has been proven that the deleted content can be reverse-engineered from the edited model \citep{patil2023can}.

\subsection{Evaluations of LLM Unlearning} 
Recent research has introduced several benchmarks and tasks for LLM unlearning from various aspects \citep{ji2024beavertails, li2024wmdp, maini2024tofu, lynch2024eight, jang2022knowledge, tian2024forget}. Among these, the Weapons of Mass Destruction Proxy (WMDP) benchmark \citep{li2024wmdp} specifically targets dangerous knowledge in biosecurity, cybersecurity, and chemical security. While these task settings primarily address the forgetting of explicit instances, they pay less attention to entity-level unlearning, which involves completely forgetting an entity. Notably, \citet{eldan2024whos} explored a particular task to unlearn the entity "Harry Potter", but it is proven that the knowledge is not entirely erased from the unlearned model \citep{shi2023detecting}. \citet{maini2024tofu} presented TOFU consisting of 200 fictitious author profiles to assess unlearning methods from Model Utility and Forget Quality. However, this benchmark only focused on ideal scenarios with exact forget sets. Similarly, the RWKU benchmark \citep{jin2024rwku} chooses 200 real-world famous people as entities for unlearning, with a more practical task setting. Nonetheless, it still focuses on specific pieces of knowledge related to target entities rather than the entire entity, making it closer to instance-level unlearning.

\section{Conclusion}\label{sec:conclu}
In this paper, we propose a novel task, entity-level unlearning for LLMs, which is required in many practical scenarios. We evaluate popular unlearning algorithms on this task and reveal that existing unlearning methods struggle to effectively erase the entire entity from the target model. Furthermore, we find that the knowledge coverage and size of forget set play crucial roles in the performance of the algorithms. Additionally, our analysis shows that entities introduced through fine-tuning are more vulnerable than pre-trained entities during unlearning, underscoring the need for more robust entity injection techniques. These findings suggest promising directions for future research, encouraging the development of more precise knowledge-probing methods and corresponding unlearning algorithms to enhance entity removal.

\section*{Limitations}\label{sec:limitations}
Despite the comprehensive analysis of entity-level unlearning task, there are several limitations worth noting. 
First, our work primarily focuses on analyzing the entity-level unlearning task and identifying potential insights for improvement, rather than proposing a specific unlearning algorithm, which will be a focus of our future research.
Secondly, the entity-level unlearning task in our work pays solely attention to single-entity deletion, omitting batch or sequential unlearning involving multiple entities, which could be further explored in future research.
Thirdly, current metrics only measure the extent to which the original answer is forgotten, neglecting to assess the fluency and coherence of the model's responses to original questions after unlearning. 
Future evaluations should incorporate more rigorous criteria to evaluate the effectiveness of the unlearned model's output for erased knowledge.

\section*{Acknowledgements}\label{sec:Acknowledgments}
Xiaocheng Feng is the corresponding author of this work. We thank the anonymous reviewers for their insightful comments. This work was supported by the National Natural Science Foundation of China (NSFC) (grant 62276078, U22B2059), the Key R\&D Program of Heilongjiang via grant 2022ZX01A32, the International Cooperation Project of PCL, PCL2022D01 and the Fundamental Research Funds for the Central Universities (Grant No.HIT.OCEF.2023018). We also thank Du Xiaoman (Beijing) Science Technology Co., Ltd for supporting part of the computing resources and funding.

\bibliography{custom}

\appendix
\section{Details for Forget Set Construction}
\label{app_framework}
Based on the unlearning framework, we develop entity-level unlearning task methods to evaluate the impact of each stage on the deletion effect. Specifically, we design a knowledge probing method that relies solely on the model, encompassing three stages:
\begin{enumerate}[leftmargin=*,itemsep=0em]
    \item First, we prompt the model to self-generate non-repetitive entity-related questions. See figure \ref{question_probing_prompt} for the prompt used.
    \item Second, we apply greedy decoding to obtain answers for each question, ensuring that the answers maintain a relatively high generation probability.
    \item Finally, recognizing that the QA pairs obtained in the first two stages may not necessarily reflect knowledge recognized by the model, we introduced a model-based self-verification process. In this stage, the model repeatedly checks each QA pair, and only those pairs consistently recognized as correct by the model are retained.
    
    To minimize the influence of positional bias on the model's responses, we randomly shuffled the positions of the "Yes" and No" options, conducted five iterations of testing, and only retained answers where the model consistently selected the "Yes" option across all five trials. See Figure \ref{QA_checking_prompt}  for the prompt used.
\end{enumerate}

\section{Experimental Details}
\subsection{Details for Target Models}
\label{target_model}
We retrain Llama2-7B-Chat \cite{touvron2023llama2} and Phi-1.5 \cite{li2023textbooks} on TOFU as the target models for entity-level unlearning following \cite{maini2024tofu} and Llama2-7b-Base \cite{touvron2023llama2} for ablation analysis with the same hyperparameters, employing AdamW with a weight decay of 0.01, a learning rate of $10^{-5}$, and a linear warmup during the first epoch. After fine-tuning,  the LLMs acquire information about the author in the TOFU, as demonstrated in table \ref{tab:finetune}.
\begin{table}[t!]
    \centering
    \begin{tabular}{lcc}
    \toprule
         Model &  Original & Finetuned on TOFU \\
         \midrule
         Llama2-7B-Chat &  0.3794 & 0.9779 \\
         Llama2-7B-Base &  - & 0.8766 \\
         Phi-1.5 &  0.4356 & 0.9232 \\
         \bottomrule
    \end{tabular}
    \caption{ROUGE scores on the TOFU dataset for Llama2-7B-Chat, Llama2-7B-Base and Phi-1.5.}
    \label{tab:finetune}
\end{table}

\subsection{Details for Evaluation Set}
\label{app_eval_set}
The evaluation set consists of the target set, the retain set, real authors set and world facts set. Each set comprises items, $S=\{(q^i,a^i,\tilde a^i, A_\text{pert}^i)\}_{i=1}^N$, each of which includes an original QA pair $\{q,a\}$, a paraphrased answer $\tilde a$ and five perturbed answers $A_\text{pert}=\{\hat a^1,...,\hat a^5\}$. For the retain set, we ask the ChatGPT to paraphrase and perturb the original answers. Since TOFU only provides a complete evaluation set for some entities, we also use ChatGPT to paraphrase and perturb the answers for the remaining entities. See figure \ref{paraphrased_answer_prompt},\ref{perturbed_answer_prompt} for the prompt used.

\subsection{Details for Baselines}
\label{app_baseline}
We evaluate five common baselines for the entity-level unlearning task, following \cite{maini2024tofu, zhang2024negative}. The specific approaches are as follows:
\begin{itemize}[noitemsep,topsep=0pt,parsep=0pt,leftmargin=*]
\item \textbf{Gradient Ascent} \citep{yao2023large}, one of the most straightforward and basic unlearning algorithms, updates the target model, which is parametrized by $\theta_t$ by maximizing the cross-entropy loss $\ell(x, \theta_t)$ over the forget set $S_F$:
\begin{equation}
L(S_F, \theta_t) =  \frac{1}{\vert S_F \vert} \sum_{x \in S_F}\ell(x, \theta_t). 
\end{equation}
\item \textbf{Gradient Difference} \citep{liu2022continual} implements unlearning on the forget set $S_F$ by gradient ascent and learning on the retain set $S_F$. The loss function we aim to minimize can be written as:
\begin{equation}
    L_{\text{GD}} = - L(S_F, \theta_t) + L(S_R,\theta_t).
\end{equation}
\item \textbf{KL Minimization}, applies a additional Kullback-Leibler (KL) divergence regularization $R_{\text{KL}}$ between the predictions on $S_R$ of the original model $\theta_t$ and the unlearned model $\theta_{t+1}$, while performing GA on $S_F$. The loss function we aim to minimize can be written as:
\begin{equation}
L_{\text{KL}} = - L(S_F, \theta_t) + R_{\text{KL}}
\end{equation}
\begin{equation}
R_{\text{KL}}= \frac{1}{\vert S_R \vert} \sum_{x \in S_R}\text{KL}\left(P(x,\theta_t) \big\Vert P(x,\theta_{t+1})\right)
\end{equation}
Where $P(x,\theta_t)$ represents a probability distribution for a sample $x \in S_F $ over the vocabulary according to the model $\theta_t$.
\item \textbf{Preference Optimization}, realigns the model $\theta_t$ to respond to the questions in the forget set $S_F$ with a refusal answer, such as "I don't know," through the DPO algorithm \citep{rafailov2024direct}, while learning on the retain set $S_R$ by gradient descent. The loss function we aim to minimize can be written as:
\begin{equation}
L_{\text{idk}} = L_{\text{DPO}}(S_F^\text{idk}, \theta_t) + L(S_R, \theta_t).
\end{equation}
Where $L_{\text{DPO}}(\cdot)$ represents the loss function of DPO \citep{rafailov2024direct}; $S_F^\text{idk}$ consists of the samples which include the original question from forget set with a refusal answer. 
\item \textbf{Negative Preference Optimization}, is inspired by DPO, which only requires the negative term during preference optimization. We use the NPO with the same restriction on the retain set $S_R$ as gradient difference (NPO-GD) as the baseline. The loss function we aim to minimize can be written as:
\begin{equation} 
L_{\text{NPO-GD}} = L_{\text{NPO}}(S_F, \theta_t) + L(S_R,\theta_t). 
\end{equation}
Where $L_{\text{NPO}}(\cdot)$ represents the loss function of basic NPO \citep{zhang2024negative};
\end{itemize}

\subsection{Details for Evaluation Metrics}
\label{app_mertic}
Following TOFU \cite{maini2024tofu}, we conduct an evaluation on the evaluation set using the following metric:
\begin{itemize}[noitemsep,topsep=0pt,parsep=0pt,leftmargin=*]
\item \textbf{ROUGE} \citep{lin2004rouge}, measures the overlap co-occurrence of n-grams between the ground truth answer and model's generation under greedy decoding for QA pairs. We reported the ROUGE-L recall score.
\item \textbf{Probability}, computes the conditional probability with length normalization $P(a|q)^{1/\vert a \vert}$ of test QA pairs $S=\{(q_i,a_i)\}_{i=1}^N$ in target set and retain set. Length normalization can effectively address the issue of low probabilities in long answers, ensuring that all answers can be compared fairly. Additionally, we calculate the normalized conditional probability of the correct answer overall answers as the final probability score on world fact set and real author set, which can be written as:
\begin{equation}
    \text{Probability} = \frac{P(a|q)^{1/\vert a \vert}}{\underset{x \in { \{a\} \cup \mathcal A_\text{pert}}}{\sum}P(x|q)^{1/\vert x \vert}}
\end{equation}
Where $a$ is the ground truth of the test question, $A_\text{pert}=\{\hat a^1,...,\hat a^5\}$ is a set consisting of the five perturbed versions of $a$.
\item \textbf{Accuracy} calculates the proportion of a paraphrased answer that the unlearned model can select from perturbed answers based on the original question. Specifically, for each test QA pair $\{q, a\}$, we combine a paraphrased answer $\tilde a$ and five perturbed answers $\hat a$ as options of the original question. The accuracy metric is defined as the proportion of paraphrases that the unlearned model $\theta_t$ can correctly identify among all test QA pairs, which can be written as:
\begin{equation}
     \text{Accuracy} = \frac{\sum_{i=1}^N D(a^i,q^i,\theta_t)}{N}
\end{equation}
\begin{equation} 
   D(a,q,\theta_t) = \mathbb{I}(\underset{x \in { \{\tilde a\} \cup \mathcal A_\text{pert}}}{\text{argmax}}{P(x|q,\theta_t)^{1/\vert x \vert}} = \tilde a)
\end{equation}
Where $\mathbb{I}(\cdot)$ is an indicator function that returns 1 if the condition is met and 0 otherwise. $P(\cdot)$ represents the conditional probability. $A_\text{pert}=\{\hat a^1,...,\hat a^5\}$ is a set consisting of the five perturbed versions of $a$.
\item \textbf{Forget Quality}, assesses the unlearning effectiveness of the unlearned model. It measures the difference between the distributions of the \textit{Truth Ratio} metric via the Kolmogorov-Smirnov (KS) test from the unlearned model and the reference model, which trained only on the $S_\text{train}=S_\text{tofu} / S_\text{entity}$. We report the $p-value$ from the KS test as the forget quality. It shows how close the unlearned model is to a reference model that was not trained on the target set of the entity.
\item \textbf{Truth Ratio} \cite{maini2024tofu}, calculates the ratio of the average probability of the perturbed versions $\hat a$ of the ground truth answer $a$ to the probability of a paraphrased version $\tilde a$ of the ground truth answer. In order to keep the score $R_\text{truth}$ between zero and one, we reported the truth ratio from $min(R_\text{truth}, 1/R_\text{truth})$. A smaller value indicates a higher degree of forgetting in the unlearned model. The $R_\text{truth}$ can be written as:
\begin{equation}
    R_\text{truth} = \frac{\frac{1}{\vert \mathcal A_\text{pert} \vert} \sum_{\hat a \in \mathcal A_\text{pert}}P(\hat a | q)^{1/\vert \hat a \vert}}{P(\tilde a | q)^{1/\vert \tilde a \vert}}
\end{equation}
Where $\mathcal A_\text{pert}$ represents a set consisting of five perturbations $\hat a$.
\end{itemize}

\begin{table*}[!ht]
\centering
\resizebox{\textwidth}{!}{
\begin{tabular}{llcccccccc}
\toprule
\multirow{2}{*}{\textbf{Method}} & \multirow{2}{*}{\textbf{Forget Set Type}} & \multicolumn{4}{c}{\textbf{Target Set}} & \multirow{2}{*}{\textbf{RS Score} {\uparr}} & \multirow{2}{*}{\textbf{RAS Score} {\uparr}} & \multirow{2}{*}{\textbf{WFS Score} {\uparr}} & \multirow{2}{*}{\textbf{Model Utility} {\uparr}} \\
\cmidrule(lr){3-6}
&  & \textbf{Prob.} {\downarr} & \textbf{ROUGE} {\downarr} & \textbf{Acc.} {\downarr} & \textbf{Forget Q.} {\uparr} & & & & \\
\midrule 
\rowcolor{gray!20}\multicolumn{10}{c}{\textbf{Phi-1.5-TOFU}} \\ 
\midrule

Original & - & 0.9271 & 0.9296 & 0.6075 & 0.0655 & 0.7010 & 0.4346 & 0.5514 & 0.5414 \\
\midrule

\multirow{2}{*}{Grad. Ascent} & target set & \textbf{0.0171} & \textbf{0.3460} & \textbf{0.2025} & \textbf{0.5242} & 0.6893 & 0.3980 & 0.5278 & \textbf{0.5121} \\
 & Con. forget set  & 0.3437 & 0.4168 & 0.5600 & 0.0626 & 0.6403 & 0.3721 & 0.5043 & 0.4814 \\
\midrule

\multirow{2}{*}{Grad. Diff.} & target set & \textbf{0.1836} & \textbf{0.4357} & \textbf{0.3650} & \textbf{0.4274} & 0.7533 & 0.4448 & 0.5662 & \textbf{0.5616} \\
 & Con. forget set  & 0.5143 & 0.4780 & 0.4925 & 0.1197 & 0.7370 & 0.4085 & 0.5447 & 0.5319 \\
\midrule

\multirow{2}{*}{KL Min.} & target set & \textbf{0.0175} & \textbf{0.3461} & \textbf{0.2025} & \textbf{0.5158} & 0.6977 & 0.3988 & 0.5285 & \textbf{0.5143} \\
 & Con. forget set  & 0.3490 & 0.4162 & 0.5600 & 0.0629 & 0.6498 & 0.3759 & 0.5096 & 0.4869 \\
\midrule

\multirow{2}{*}{Pref. Opt.} & target set & \textbf{0.4323} & \textbf{0.1979} & \textbf{0.5050} & \textbf{0.1584} & 0.8118 & 0.3883 & 0.5673 & 0.5386 \\
 & Con. forget set  & 0.4811 & 0.2303 & 0.5225 & 0.1414 & 0.8111 & 0.4028 & 0.5615 & \textbf{0.5458} \\
\midrule

\multirow{2}{*}{NPO-GD} & target set & \textbf{0.0749} & \textbf{0.4016} & \textbf{0.3300} & \textbf{0.5104} & 0.7674 & 0.4246 & 0.5477 & \textbf{0.5471} \\
 & Con. forget set  & 0.7567 & 0.6873 & 0.6000 & 0.0794 & 0.7682 & 0.4225 & 0.5451 & 0.5451 \\

\bottomrule 
\end{tabular}
}
\caption{
The performance of the Phi-1.5-TOFU after entity-level unlearning under the five algorithms. We list the results when the Forget Quality (Forget Q.) reaches the first peak score. Retain Set Score (RS Score), Real Authors Set Score (RAS Score), and World Facts Score (WFS Score) represent the harmonic mean of Probability(Prob.), ROUGE, and Accuracy(Acc.) on their respective sets. \uparr \ \ represents that the higher score is better, while \downarr \ \ indicates the opposite. The results in \textbf{bold} represent the best results between the two forget sets.
}
\label{table:main-phi_results}
\end{table*}

\begin{table*}[!ht]
\centering
\resizebox{\textwidth}{!}{
\begin{tabular}{llcccccccccc}
\toprule
&
& \multicolumn{3}{c}{\textbf{Retain Set}} 
& \multicolumn{3}{c}{\textbf{Real Authors Set }} 
& \multicolumn{3}{c}{\textbf{World Facts Set}} \\

\cmidrule(lr){3-5}
\cmidrule(lr){6-8}
\cmidrule(lr){9-11}

\multirow{-2}{*}{\textbf{Method}} & \multirow{-2}{*}{\textbf{Forget Set type}}
& \textbf{Probability} {\uparr} & \textbf{ROUGE} {\uparr} & \textbf{Accuracy} {\uparr}
& \textbf{Probability} {\uparr} & \textbf{ROUGE} {\uparr} & \textbf{Accuracy} {\uparr}
& \textbf{Probability} {\uparr} & \textbf{ROUGE} {\uparr} & \textbf{Accuracy} {\uparr}
\\

\midrule 
\rowcolor{gray!20} \multicolumn{11}{c}{\textbf{Llama2-7B-chat-TOFU}} \\ 
\midrule

Finetuned & - & 0.8724 & 0.8597 & 0.8894 & 0.4478 & 0.9330 & 0.5600 & 0.4251 & 0.8832 & 0.4615 \\
\midrule

\multirow{2}{*}{Grad. Ascent} & target set & 0.5856 & 0.7082 & 0.7752 & 0.4197 & 0.8786 & 0.5260 & 0.3956 & 0.8600 & 0.4376  \\
 & Con. forget set  & 0.3982 & 0.7169 & 0.7295 & 0.4218 & 0.8747 & 0.4875 & 0.3840 & 0.8600 & 0.4115  \\
\midrule

\multirow{2}{*}{Grad. Diff.} & target set & 0.8397 & 0.8124 & 0.8464 & 0.4789 & 0.9073 & 0.7320 & 0.4656 & 0.8918 & 0.5504 \\
 & Con. forget set  & 0.7383 & 0.8059 & 0.8128 & 0.3789 & 0.8996 & 0.4215 & 0.3674 & 0.8738 & 0.3996  \\
\midrule

\multirow{2}{*}{KL Min.} & target set & 0.4854 & 0.6947 & 0.7483 & 0.4317 & 0.7790 & 0.5245 & 0.4019 & 0.8602 & 0.4491  \\
 & Con. forget set & 0.4109 & 0.7177 & 0.7265 & 0.4241 & 0.8766 & 0.4960 & 0.3844 & 0.8598 & 0.4111  \\
\midrule

\multirow{2}{*}{Pref. Opt.} & target set & 0.9308 & 0.9239 & 0.8564 & 0.5075 & 0.7770 & 0.8550 & 0.4723 & 0.8980 & 0.6692  \\
 & Con. forget set & 0.9322 & 0.9265 & 0.8585 & 0.5120 & 0.8565 & 0.8520 & 0.4730 & 0.9054 & 0.6714 \\
 \midrule

\multirow{2}{*}{NPO-GD} & target set & 0.7701 & 0.7740 & 0.8244 & 0.4252 & 0.9004 & 0.5595 & 0.4157 & 0.8767 & 0.4697  \\
 & Con. forget set & 0.6715 & 0.7694 & 0.8126 & 0.3515 & 0.9111 & 0.3765 & 0.3537 & 0.8777 & 0.3949  \\

\midrule 
\rowcolor{gray!20} \multicolumn{11}{c}{\textbf{Phi1.5-TOFU}} \\ 
\midrule

Finetuned & - & 0.6107 & 0.7448 & 0.7696 & 0.3780 & 0.4157 & 0.5400 & 0.4088 & 0.7560 & 0.5983 \\
\midrule

\multirow{2}{*}{Grad. Ascent} & target set & 0.7206 & 0.7825 & 0.5929 & 0.3821 & 0.3254 & 0.5415 & 0.3969 & 0.7186 & 0.5641 \\
 & Con. forget set  & 0.6061 & 0.7472 & 0.5893 & 0.3740 & 0.2844 & 0.5340 & 0.3843 & 0.6787 & 0.5338 \\
\midrule

\multirow{2}{*}{Grad. Diff.} & target set & 0.8623 & 0.8422 & 0.6115 & 0.3938 & 0.4101 & 0.5660 & 0.4165 & 0.7897 & 0.6132 \\
 & Con. forget set  & 0.8432 & 0.8442 & 0.5882 & 0.3881 & 0.3371 & 0.5555 & 0.4052 & 0.7339 & 0.5962 \\
\midrule

\multirow{2}{*}{KL Min.} & target set & 0.7394 & 0.7882 & 0.5957 & 0.3826 & 0.3264 & 0.5420 & 0.3977 & 0.7217 & 0.5628  \\
 & Con. forget set  & 0.6299 & 0.7541 & 0.5872 & 0.3751 & 0.2900 & 0.5360 & 0.3860 & 0.6838 & 0.5453 \\
\midrule

\multirow{2}{*}{Pref. Opt.} & target set & 0.8895 & 0.9181 & 0.6748 & 0.3560 & 0.3330 & 0.5225 & 0.4051 & 0.7797 & 0.6504 \\
 & Con. forget set  & 0.8879 & 0.9127 & 0.6771 & 0.3493 & 0.3779 & 0.5160 & 0.4016 & 0.7747 & 0.6402 \\
 \midrule

\multirow{2}{*}{NPO-GD} & target set & 0.8331 & 0.8708 & 0.6408 & 0.3798 & 0.3863 & 0.5425 & 0.4059 & 0.7545 & 0.5923  \\
 & Con. forget set & 0.7992 & 0.8819 & 0.6579 & 0.3745 & 0.3933 & 0.5295 & 0.4017 & 0.7578 & 0.5902  \\

\bottomrule 
\end{tabular}
}
\caption{
The performance of the fine-tuned LLMs after entity-level unlearning under Gradient Ascent (Grad. Ascent), Gradient Difference (Grad. Diff.), KL Minimization (KL Min.), Preference Optimization (Pref. Opt.), and Negative Preference Optimization with gradient descent on the retain set (NPO-GD) on the retain set, real authors set and world facts set.
}
\label{table:retain_sets_full}
\end{table*}

\begin{table}[ht]
\centering
\resizebox{\columnwidth}{!}{
\begin{tabular}{llccc}
\toprule
& \textbf{Forget Set type} & \textbf{Probability} {\uparr} & \textbf{ROUGE} {\uparr} & \textbf{Accuracy} {\uparr}
\\

\midrule 
\rowcolor{gray!20} \multicolumn{5}{c}{\textbf{Llama2-7B-TOFU}} \\ 
\midrule

Finetuned & - & 0.9908 & 0.9839 & 0.6297 \\
\midrule

\multirow{2}{*}{Grad. Ascent} & target set & 0.7829 & 0.6444 & 0.5635 \\
 & Con. forget set  & 0.4774 & 0.5672 & 0.5637 \\
\midrule

\multirow{2}{*}{Grad. Diff.} & target set & 0.9444 & 0.8160 & 0.5895 \\
 & Con. forget set  & 0.7139 & 0.6477 & 0.5580 \\
\midrule

\multirow{2}{*}{KL Min.} & target set & 0.4577 & 0.4425 & 0.5270 \\
 & Con. forget set & 0.4853 & 0.5702 & 0.5653 \\
\midrule

\multirow{2}{*}{Pref. Opt.} & target set & 0.7224 & 0.0801 & 0.5937 \\
 & Con. forget set & 0.7339 & 0.1398 & 0.5928 \\
 \midrule

\multirow{2}{*}{NPO-GD} & target set & 0.8797 & 0.7301 & 0.5838 \\
 & Con. forget set & 0.8599 & 0.7585 & 0.5928 \\

\midrule 
\rowcolor{gray!20} \multicolumn{5}{c}{\textbf{Phi1.5-TOFU}} \\ 
\midrule

Finetuned & - & 0.9272 & 0.9154 & 0.6130 \\
\midrule

\multirow{2}{*}{Grad. Ascent} & target set & 0.6742 & 0.6108 & 0.5677 \\
 & Con. forget set  & 0.6297 & 0.5836 & 0.6012 \\
\midrule

\multirow{2}{*}{Grad. Diff.} & target set & 0.8154 & 0.6841 & 0.5498 \\
 & Con. forget set  & 0.7759 & 0.6589 & 0.5623 \\
\midrule

\multirow{2}{*}{KL Min.} & target set & 0.6859 & 0.6130 & 0.5707 \\
 & Con. forget set  & 0.6397 & 0.5861 & 0.6005 \\
\midrule

\multirow{2}{*}{Pref. Opt.} & target set & 0.7604 & 0.3400 & 0.5637 \\
 & Con. forget set  & 0.7779 & 0.4057 & 0.5650 \\
 \midrule

\multirow{2}{*}{NPO-GD} & target set & 0.8442 & 0.7628 & 0.6107 \\
 & Con. forget set & 0.8875 & 0.8726 & 0.6373 \\

\bottomrule 
\end{tabular}
}
\caption{
The performance of the fine-tuned LLMs after entity-level unlearning under Gradient Ascent (Grad. Ascent), Gradient Difference (Grad. Diff.), KL Minimization (KL Min.), Preference Optimization (Pref. Opt.), and Negative Preference Optimization with gradient descent on the retain set (NPO-GD) on other entities.
}
\label{table:other_entities_full}
\end{table}
\subsection{Details for Experiments}
\label{experiments_results}
We applied all five unlearning algorithms to the forget set, utilizing the AdamW with warm-up during the first epoch, a batch size of 4, and a learning rate of $10^{-5}$. The evaluation was carried out on the target models: Llama2-7B-Chat-TOFU and Phi-1.5-TOFU. Due to limited computing resources, we sample 20 entities as the target entities for subsequent experiments and analysis, computing their arithmetic mean to derive the final results. Our main experiment and analysis were conducted on 8x80GB A100 GPUs for approximately one week.

The main experimental results of Phi-1.5-TOFU are presented in table \ref{table:main-phi_results}, revealing a phenomenon analogous to that observed with Llama2-7B-chat-TOFU. Furthermore, the comprehensive experimental results for \textit{retain set}, \textit{real authors set}, and \textit{world facts set} are shown in table \ref{table:retain_sets_full}. We also assess the impact of other non-deleted entities during unlearning, as shown in table \ref{table:other_entities_full}.

\section{Details for Analysis}
\label{analysis}
\subsection{Analytical Experiments on Forget Set}
\label{forget_set_appendix}
When analyzing the impact of forget sets on the unlearning effect, we introduce a novel metric, \textbf{Knowledge Coverage}, designed to assess the overlap in knowledge between a forget set and a target set. Specifically, for each QA pair in the forget set, we compute the BERTScore against every QA pair in the target set, retaining the highest score and its corresponding target set index. It is possible for QA pairs in the target set to achieve multiple BERTScores, meaning they may be the best match for several QA pairs in the forget set. Knowledge Coverage is then calculated as the ratio of the sum of the maximum BERTScores for each sample in the target set to the total number of QA pairs in the target set. The $KC_{score}$ can be written as:
\begin{equation}
   KC_{score} = \frac{\sum_{j=1}^{\vert S_T \vert} M(j)}{\vert S_T \vert}
\end{equation}
\begin{equation}
    M(k) = \text{max}{\ B(S_F^i, S_T^k)},\text{s.t.}\ I_{max}(i)=k
\end{equation}
\begin{equation}
    I_{max}(i) = \underset{j}{\text{argmax}}{\ B(S_F^i, S_T^j)}   
\end{equation}
Where $B(\cdot)$ is the function to calculate the BERTScores between the forget set $S_F$ and the target set $S_T$.

For all five algorithms, we conduct analytical experiments on the size of the constructed forget set, using AdamW with warm-up during the first epoch, a batch size of 2, a learning rate of $10^{-5}$, and 6 epochs.

Additionally, for all five algorithms, we also conduct analytical experiments on the Knowledge Coverage of the constructed forget set, using AdamW with warm-up during the first epoch, a batch size of 4, a learning rate of $10^{-5}$, and 12 epochs. Specially, we create multiple constructed forget sets by substituting different ratios of QA pairs from the target set, while keeping the total size constant at 20, to examine the impact on Knowledge Coverage. The Knowledge Coverage of each constructed forget set is shown in Table \ref{tab:forget_set_quality}.

\begin{table}[t]
\small
\centering
\resizebox{\columnwidth}{!}{
\begin{tabular}{lc}
\toprule
\textbf{Forget Set type} & \textbf{Knowledge Coverage} \\
\midrule
original Con. forget set &  0.3144 \\
\qquad w/ \textit{25\% facts replaced} &  0.4477 \\
\qquad w/ \textit{50\% facts replaced} &  0.5994 \\
\qquad w/ \textit{75\% facts replaced} &  0.7861 \\
\qquad w/ \textit{100\% facts replaced} &  1.0000 \\
\bottomrule
\end{tabular}
}
\caption{Knowledge Coverage of the constructed forget sets obtained by replacing different percentages of QA pairs from the target set.}
\label{tab:forget_set_quality}
\end{table}

\subsection{Analytical Experiments on Unlearning Steps}
\label{ablation_steps}
\begin{figure*}[h]
\centering
\includegraphics[width=1.0\textwidth]{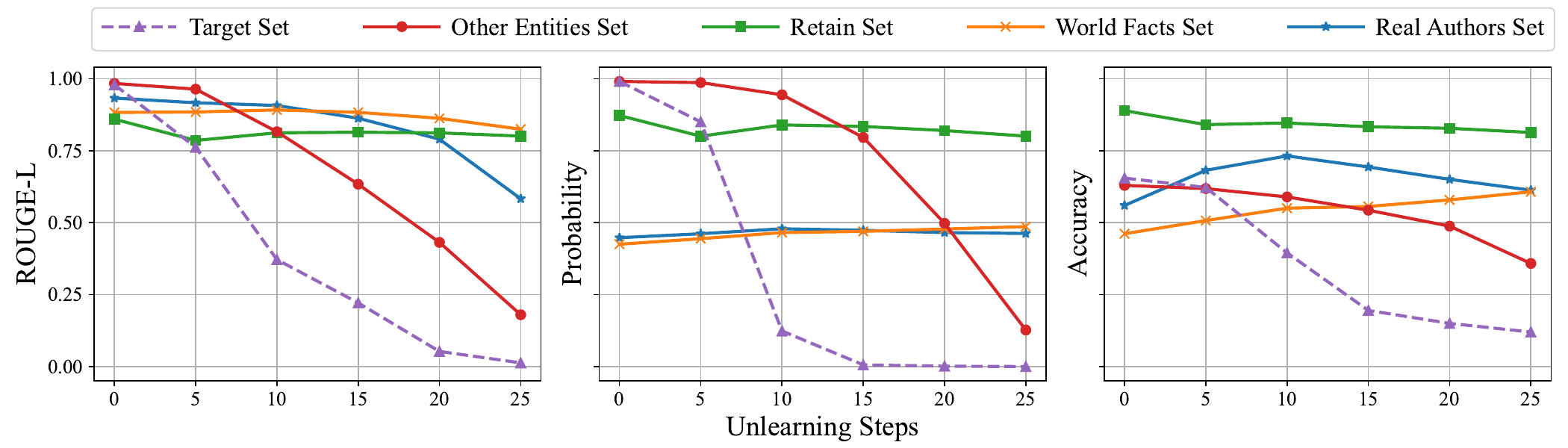} 
\caption{Step ablation analysis of unlearning Llama2-7B-Chat-TOFU using Grad. Diff. averaged over 20 entities. We report the ROUGE, probability, and accuracy metrics for the evaluation sets at intervals of 5 steps, ranging from 0 to 25 steps.}
\label{llama2_step_GD}
\end{figure*}

\begin{figure*}[h]
\centering
\includegraphics[width=1.0\textwidth]{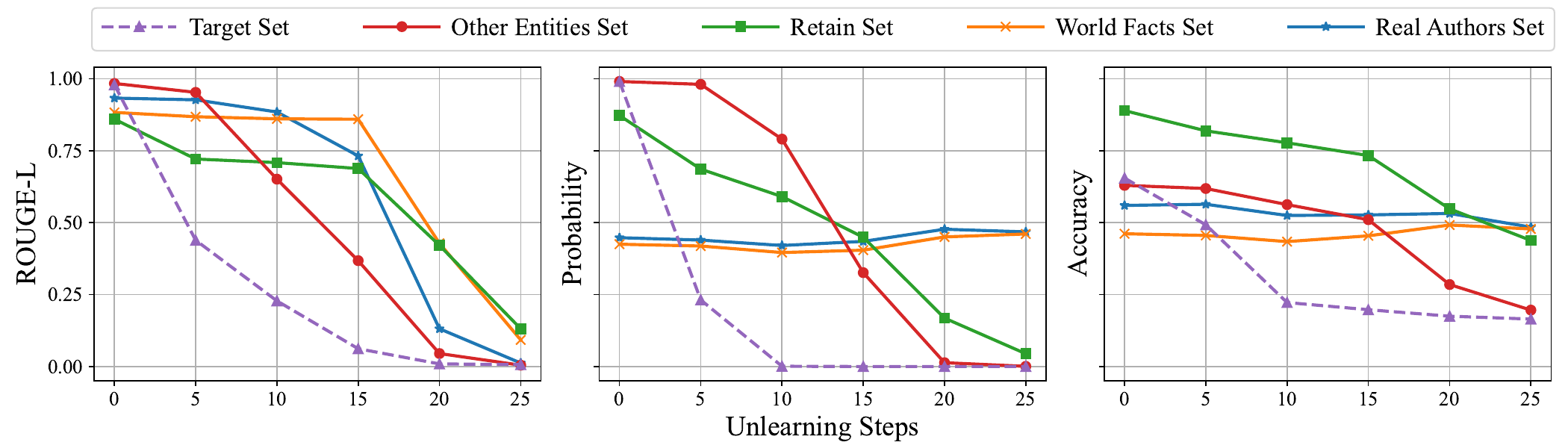} 
\caption{Step ablation analysis of unlearning Llama2-7B-Chat-TOFU using KL Min. averaged over 20 entities. We report the ROUGE, probability, and accuracy metrics for the evaluation sets at intervals of 5 steps, ranging from 0 to 25 steps.}
\label{llama2_step_kl}
\end{figure*}

\begin{figure*}[h]
\centering
\includegraphics[width=1.0\textwidth]{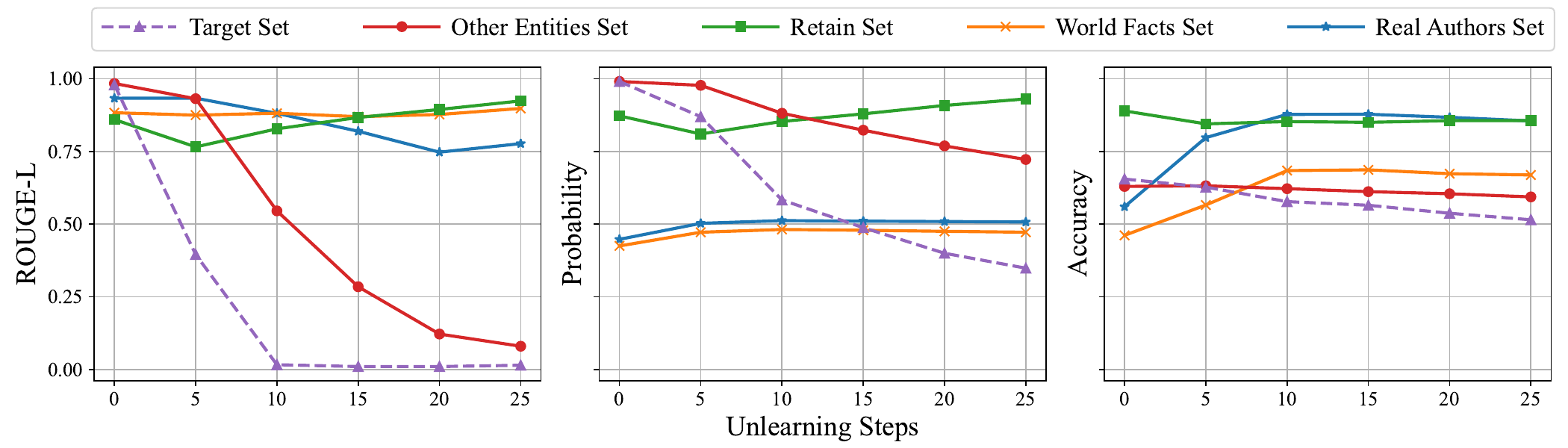}
\caption{Step ablation analysis of unlearning Llama2-7B-Chat-TOFU using the Pref. Opt. averaged over 20 entities. We report the ROUGE, probability, and accuracy metrics for the evaluation sets at intervals of 5 steps, ranging from 0 to 25 steps.}
\label{llama2_step_idk}
\end{figure*}

\begin{figure*}[h]
\centering
\includegraphics[width=1.0\textwidth]{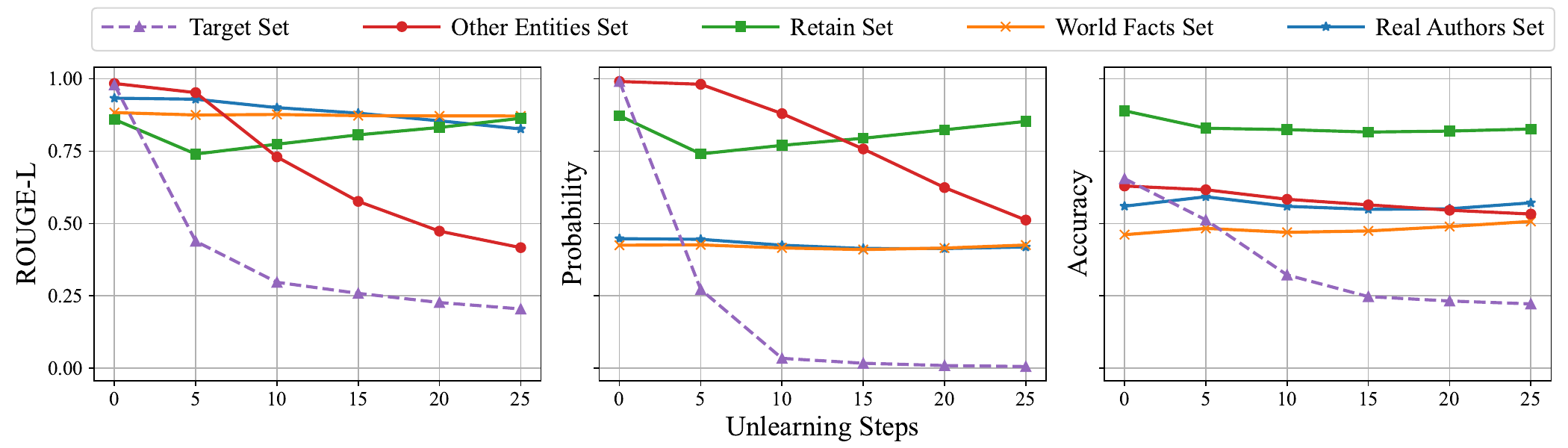}
\caption{Step ablation analysis of unlearning Llama2-7B-Chat-TOFU using NPO-GD averaged over 20 entities. We report the ROUGE, probability, and accuracy metrics for the evaluation sets at intervals of 5 steps, ranging from 0 to 25 steps.}
\label{llama2_step_npo_GD}
\end{figure*}

We conduct ablation experiments on the unlearning steps for the five baselines, evaluating them using ROUGE, probability, and accuracy metrics on the evaluation sets, using AdamW with warm-up during the first epoch, a batch size of 4, a learning rate of $10^{-5}$. There are results shown Gr for Grad. Ascent (Figure \ref{llama2_step_GA}), Grad. Diff. (Figure \ref{llama2_step_GD}), KL Min. (Figure \ref{llama2_step_kl}), Pref. Opt. (Figure \ref{llama2_step_idk}) and NPO-GD (Figure \ref{llama2_step_npo_GD}).

\subsection{Analytical Experiments on Pre-training Knowledge and Fine-tuning Knowledge}
\label{pretraining_fintuning}
We compare entity-level unlearning tasks for fine-tuned entities and pre-trained entities using the Llama2-7B-Base model. Fine-tuned entities are introduced from the TOFU dataset, resulting in the target model Llama2-7B-Base-TOFU as shown in Table \ref{tab:finetune}, from which 20 entities were sampled to form the target sets. For the pre-trained entities, we extract the factual QA pairs of celebrities based on Wikipedia using ChatGPT, retaining only those QA pairs that the model can answer correctly under greedy decoding, following \citet{gekhman2024does}. Ultimately, this process yields 20 pre-trained entities, each also containing 20 entity-related QA pairs, constituting the target sets. Then, we perform single-entity unlearning on 20 entities across two scenarios. We designate the target set as the forget set and calculate the average of their indicators as the final result to mitigate the impact of outliers.

\section{Prompts}
\label{prompt}
\begin{figure*}[h]
\centering
\includegraphics[width=1.0\textwidth]{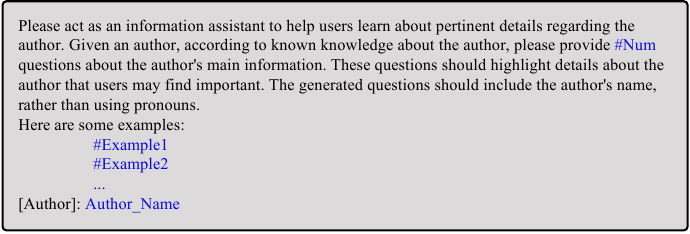} 
\vspace{-8mm}
\caption{Prompt used to generate questions based on authors' names. Specifically, we aim to prompt the target model to generate multiple questions based on \textcolor{blue}{Author\_Name}, using few-shot in-context-learning.}
\label{question_probing_prompt}
\end{figure*}

\begin{figure*}[h]
\centering
\includegraphics[width=1.0\textwidth]{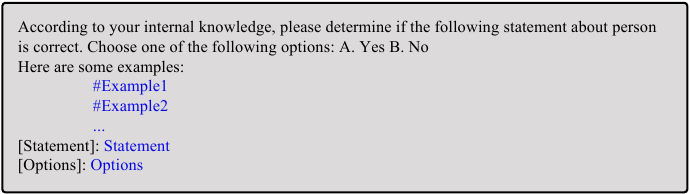} 
\vspace{-8mm}
\caption{Prompt used to check the QA pairs. Specifically, we aim to prompt the target model to self-check their own generated QA pairs based on \textcolor{blue}{Statement}, using few-shot in-context-learning.}
\label{QA_checking_prompt}
\end{figure*}

\begin{figure*}[h]
\centering
\includegraphics[width=1.0\textwidth]{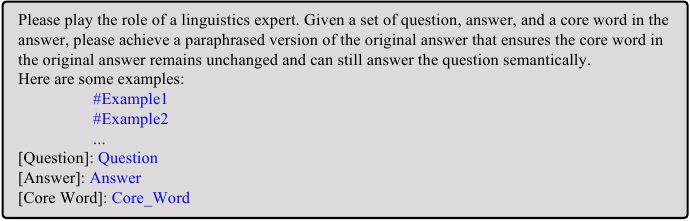} 
\vspace{-8mm}
\caption{Prompt used to paraphrase an answer based on question and core word. Specifically, we aim to prompt ChatGPT/GPT-4 to generate a paraphrased version of \textcolor{blue}{Answer} based on \textcolor{blue}{Question} and \textcolor{blue}{Core\_Word}, using few-shot in-context-learning.}
\label{paraphrased_answer_prompt}
\end{figure*}

\begin{figure*}[h]
\centering
\includegraphics[width=1.0\textwidth]{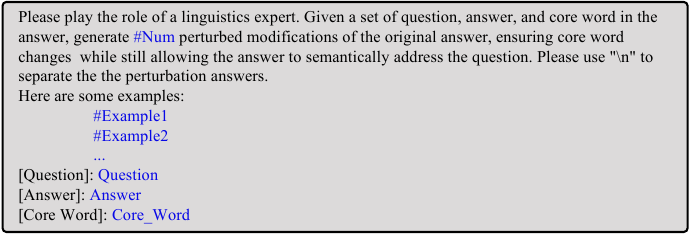} 
\vspace{-8mm}
\caption{Prompt used to perturb an answer based on question and core word. Specifically, we aim to prompt ChatGPT/GPT-4 to generate five perturbed modifications of \textcolor{blue}{Answer} based on \textcolor{blue}{Question} and \textcolor{blue}{Core\_Word}, using few-shot in-context-learning.}
\label{perturbed_answer_prompt}
\end{figure*}

\begin{figure*}[h]
\centering
\includegraphics[width=1.0\textwidth]{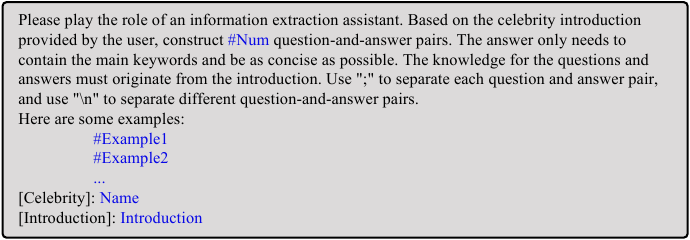} 
\vspace{-8mm}
\caption{Prompt used to extract question-answer pairs from an introduction. Specifically, we aim to prompt ChatGPT/GPT-4 to extract question-answer pairs based on \textcolor{blue}{Name} and \textcolor{blue}{Introduction}, using few-shot in-context-learning.}
\label{QA_pairs_extract_prompt}
\end{figure*}

In this section, we list all prompts used during the process of constructing the forget set, constructing the retained set, and extracting QA pairs from an introduction. which include questions probing (Figure \ref{question_probing_prompt}), QA checking (Figure \ref{QA_checking_prompt}), answers paraphrasing (Figure \ref{paraphrased_answer_prompt}), answers perturbing (Figure \ref{perturbed_answer_prompt}) and QA pairs extraction from an introduction (Figure \ref{QA_pairs_extract_prompt}).

\end{document}